# Utility of General and Specific Word Embeddings for Classifying Translational Stages of Research.

Vincent Major, ME MS[1], Alisa Surkis, PhD[1], and Yindalon Aphinyanaphongs, MD PhD[1]
[1]NYU Langone Health, New York, NY, USA

**Abstract**

*Conventional text classification models make a bag-of-words assumption reducing text into word occurrence counts per document. Recent algorithms such as word2vec are capable of learning semantic meaning and similarity between words in an entirely unsupervised manner using a contextual window and doing so much faster than previous methods. Each word is projected into vector space such that similar meaning words such as "strong" and "powerful" are projected into the same general Euclidean space. Open questions about these embeddings include their utility across classification tasks and the optimal properties and source of documents to construct broadly functional embeddings. In this work, we demonstrate the usefulness of pre-trained embeddings for classification in our task and demonstrate that custom word embeddings, built in the domain and for the tasks, can improve performance over word embeddings learnt on more general data including news articles or Wikipedia.*

**Introduction**

Vector representations of language have become very popular in recent years due to their ability to recover semantic and syntactic information about natural language. Popular approaches consume vast corpora of text and train in hours or days rather than weeks or months. However, the performance improvement of using a relatively small domain-specific corpus to create custom embeddings, compared to a large, diverse corpus of general text remains unknown. If an improvement in classification performance, is possible, customization of vector representations for the specific task would be warranted. However if not, time and resources could be saved by using a previously validated set of embeddings.

The Clinical and Translational Science Award (CTSA) program was started in 2006 by the National Institutes of Health and includes over sixty academic medical institutions and their partners across the United States. The program aims to: "transform the local, regional, and national environment to increase the efficiency and speed of clinical and translational research across the country"[1]. Evaluation of the CTSA program's effectiveness is hampered by the lack of a method for measuring the impact of the program's support on the speed of translation. The first step towards development of such a method is to clearly define the phases of the translational research spectrum, the second is to operationalize application of those definitions.

Translational biomedical research can be conceptualized along a spectrum beginning with basic sciences, traversing through preclinical and human studies then progressing towards population level studies. Recent work by Surkis et al.[2] attempted to add rigor to the classification process by creating a checklist to guide the manual process of labeling publications within the five stages of translational research, T0 to T4. The authors applied the checklist to MEDLINE indexed publications published by CTSA affiliated institutions to construct a set of 386 coded abstracts.

Manual coding and review is an immensely laborious task; Surkis et al.[2] attempted to operationalize their system with a supervised machine learning approach. Several different model types were employed to classify publications into their translational class. The training sets combined T1 and T2 categories into T1/T2, and similarly for T3 and T4 into T3/T4, due to low frequency of these publication types compared to T0, reducing the task to three-class classification. Surkis and colleagues benchmarked each algorithm reporting predictive performance, measured by the mean area under the receiver operating characteristic curve (AUC), ranging from 0.84-0.94 within 5-fold cross validation experiments.

The algorithms employed in the original work make a common bag-of-words assumption and use standard term frequency–inverse document frequency (tf–idf) token encodings. The limitations of this traditional approach are well known and include loss of word order and loss of semantic equivalence in that tokens such as "strong" and "powerful" are considered distinct and a model will learn weightings independently on a very large vocabulary[3]. The latter is mediated by large training datasets where both words are expected to occur frequently enough for a reliable model coefficient. More recently, focus has shifted to text classifications algorithms, in particular neural networks[4], that do not make the bag-of-words assumption globally, instead observing local context windows, and can learn text representations, in an entirely unsupervised manner, that capture semantics and can be applied to a variety of tasks.

These unsupervised techniques represent text as vectors in a dense, N-dimensional space. One very popular method, *word2vec*, uses a sliding context window, to observe how similar words are used in similar contexts. This procedure projects similar words into similar vector space and thus learning semantic meaning and similarity of words[5,6]. Representing words in vector space, constructed using only the context around each word, allows *word2vec* to represent millions of unique words in hundreds of dimensions. Each pair of words are related by their geometric distance in space, a more powerful representation than if the word exists or not. For example, the vectors for "strong" and "powerful" will be close together in Euclidian space, and far from an unrelated word such as "tomorrow". Other mathematical operations are also possible between words[5], for example "Paris - France + Italy = Rome". *word2vec* has proven to be a very successful algorithm for unsupervised learning of semantic similarity and relatedness, including in biomedical tasks[7–9] and compared against another competing, non-neural network, algorithm: *GloVe*[9,10].

Aside from optimizing semantic similarity and relatedness, *word2vec* offers advantages for word representation in classification, over many alternative approaches, by 1) mediating the deleterious effects of the classical bag-of-words assumption, 2) representing words as vectors rather than binary occurrences, and 3) learning semantic meaning of words, in an unsupervised manner, using a corpus much larger than can be feasibly labelled manually, and 4) allowing fine-tuning of the embedding space in downstream models. Exploiting word2vec embeddings can lead to a more generalizable text classification model (a semi-supervised approach[4]).

However, intrinsic evaluation measures, including semantic similarity and relatedness, do not necessarily translate to downstream extrinsic, domain- or task-specific performance. It is possible that *word2vec* could align, in vector space, a set of related words—that is, a set of words that are frequently used in similar contexts—despite their differing semantic meaning. An example[11] could be emotions such as "*angry*", "*sad*", "*happy*" that can all be used in similar sentences: "*I felt <emotion>.*" or "*They were feeling <emotion> because...*" etc. An unsupervised method such as *word2vec* could naively cluster these, quite disparate, emotions together due to their shared context, in this case some form of "*feel*". Similarly in translational research, similar concepts should group together for example, medications such as statins to treat cardiovascular disease, or proteins associated with transcription factors. Intrinsic evaluation may not penalize their closeness (sufficiently) whereas an extrinsic evaluation in a task where emotion is important, for example sentiment analysis, would suffer from their closeness in vector space (i.e. a downstream classifier would likely struggle to learn a separating cutting plane, or equivalent). However, an unrelated extrinsic evaluation task would likely not suffer from the cluster of emotions describing words. In other words, intrinsic evaluation measures are global but external evaluation measures are local and task-specific.

The bottom line is that pre-trained, intrinsically evaluated embeddings should better generalize semantics of natural language with the possible limitation of extrinsic performance. Overfitting to the external task is much less likely with pre-trained embeddings but performance may suffer[12]. An analyst's choice between pre-trained and custom embeddings will likely depend on the size of the available data. We present empirical results comparing several well-known pre-trained embeddings to several sets learnt in a moderately sized dataset of biomedical literature abstracts in their ability to correctly classify abstracts into their translational stage.

In a typical machine learning framework, model parameters are optimized for the task. For example, support vector machines are optimized for cost and regularized logistic regressions are optimized for the regularization parameter. *word2vec* has multiple parameters to optimize but no implicit goodness-of-fit metric (typically internal semantic similarity tasks are used). The first two parameters to consider are 1) the model: either continuous bag-of-words (CBOW) or skip-gram, and 2) the optimization technique: either hierarchical softmax or negative sampling. Both model options have their advantages with CBOW, which averages the context vectors, being faster and generally more accurate for common words in contrast to skip-gram, which does not average the context vectors, tends to be superior in smaller datasets and at representing rare words[6]. The two optimization technique alternatives are also very different where hierarchical softmax is normalized and behaves in a tree-like manner to avoid the very large vocabulary size, whereas negative sampling is un-normalized and works by randomly sampling a small number of words. The latter is quicker (2-10x) and performs especially well representing frequent words[6].

Since *word2vec* represents words, and their semantics, as vectors, the user must choose a parameter setting for both the size of the context window (typically ~5) and the number of dimensions (typically 100-300)[6,12,13]. In theory, incorporating more context and representing words in more dimensions could incorporate more information and thus improve predictive performance (while sacrificing some computational efficiency). This hypothesis is supported by increasing semantic similarity and relatedness on a benchmark biomedical concept task [7]. Previous work on our classification task, trialed both model options (CBOW vs skip-gram), both optimization techniques (hierarchical softmax and negative sampling), and a simple grid search of dimension sizes (50, 100, 150, 200, and 300) and context window lengths (4, 5, 6, 8, and 10) to assess the sensitivity of *word2vec* to parameter selection.

Vector representations of text can recover semantic and syntactic information about natural language. However, it remains unknown if these tools are broadly applicable to domain-specific tasks such as text classification and if they improve on the current state of the art. This paper makes several contributions.

(1) Introduces a small sample task of classifying translation research.
(2) Evaluates the effectiveness of domain-specific word embeddings compared to publicly available embeddings.
(3) Introduces an initial analysis of errors made by each classifier.

**Methods**

*Dataset*

Since publication of Surkis et al.[2], 156 publications have been added to the dataset, yielding a total of 542 MEDLINE abstracts labeled as one of three categories: T0 (n = 281), T1/T2 (n = 109), and T3/T4 (n = 152). The labels correspond to descriptions of the stage of translational research where T0 is fundamental science and T3/T4 are larger late-stage clinical trials and population level studies where, for this work, the three classes can be considered mutually exclusive. The class imbalance is biased as the majority, T0, group was undersampled during manual labelling to preferentially improve sample size across the other classes given limited resources. After the initial random sampling produced a very small sample of T1/T2 articles, additional sets of articles to be labeled were selected by filtering the search with a set of terms that returned results with an increased proportion of T1/T2 articles, as described in Surkis et al.[2].

*Word Embeddings*

(a) General Word Embeddings
Learning domain-specific embeddings requires a sufficiently large corpus and computational resources. In the absence of the necessary data or resources, embeddings that have been previously validated and published as part of a prior work may prove adequate for many tasks despite being created from general, non-domain-specific, text. To assess utility of general word embeddings, we used publicly available embeddings that can be used 'off-the-shelf'. Five sets of word embeddings were found ranging from general text corpora utilizing news articles or Wikipedia to more domain specific using Pubmed and PubMed Central (PMC). The exact details of the corpora, their preprocessing (for example, punctuation/accents, grammatical case, digits, stemming etc.) and the construction of vector representations are not completely known. Each set has been validated in some internal task and published alongside an academic publication[5,14,15].

(b) Specific Word Embeddings
(i) *Data:* We constructed a domain-specific corpus of biomedical documents to allow *word2vec* to learn word embeddings specific to biomedical research. We selected all MEDLINE indexed publications, with complete titles and abstracts, published between January 2000 to December 2016, resulting in 10.5 million documents. We selected this timeframe to extend ~5 years before the start of the CTSA program (2006).

We processed the raw data by extracting and combining the title and abstract fields from each article and removing all punctuation, as recommended by the creators [16], resulting in a corpus containing 1.67 billion words, and 822,000 unique words. Very common, or stop, words are not removed as these words are often domain specific and arbitrarily selected.

(ii) *Parameter Selection:* Previous work on this task explored predictive performance over a range of *word2vec* parameters[17]. The results showed that 1) skip-gram outperforms CBOW and is less sensitive to other parameters but is slower (~4x), 2) hierarchical softmax slightly outperforms negative sampling but is slower (~4x), and 3) high or low dimension and context lengths degrade performance with the maximum observed at dimension = 200 and context = 5. Naturally, these results are specific to our PubMed corpus and classification task however, they are consistent with the literature[5–8].

- (c) *Word Embeddings Approaches*
  - (i) *word2vec*
    Rather than the bag-of-words assumption, where each unique word token is weighted for its occurrence, *word2vec* learns word embeddings in an unsupervised manner. Word embeddings, or vectors, encode unique tokens as dense N dimensional vectors, frequently 100-1000. The mapping of each word into N dimensions allows semantically similar words or inflectional variations to overlay without ontologies or lemmatization. *word2vec* consists of a relatively simple model that operates in an unsupervised manner, such that very large corpora can be used to learn embeddings much faster than previous algorithms.

  - (ii) *fastText*
    Another state of the art algorithm called *fastText*[18] has advanced the success of *word2vec*. *fastText* learns word embeddings in a manner very similar to *word2vec* except *fastText* enriches word vectors with subword information using character n-grams of variable length[14]. These character n-grams allow the algorithm to identify prefixes, suffixes, stems, and other phonological, morphological and syntactic structure in a manner that does not rely on words being used in similar context and thus being represented in similar vector space. The authors claim that their embeddings are superior to *word2vec*'s since they can be averaged to create meaningful phrase and sentence embeddings[18].

    To explore differences in algorithms, three sets of embeddings were created with identical model parameters: 1) *word2vec* CBOW, 2) *word2vec* skip-gram, and 3) *fastText* skip-gram. We hypothesize that skip-gram will outperform CBOW, as originally described[5], and *fastText* will outperform *word2vec*, due to the additional character level information. The default *word2vec* and *fastText* parameters were used, except those explicitly mentioned, including the *fastText* minimum and maximum character n-gram parameters values of 3 and 6.

*Classification Algorithms*

- (a) *Baseline Models*
  We employed four baseline models: naive Bayes (*MALLET* v2.0.7[19]), Bayesian logistic regression (*BBRBMR* v3.0[20]), support vector machines (*LIBLINEAR* v1.96[21]), and random forest (*FEST*[22]). All four algorithms use the very common bag-of-words assumption which disregards local semantic context. All four implementations use term frequency–inverse document frequency token encodings and thus require features to be created using the entire corpus (i.e. combined training, validation and testing sets) to ensure all unique words have been observed. Preprocessing is identical to the previous work[2] and consistent across models.

- (b) *fastText*
  Whereas *word2vec* cannot classify text, *fastText* can. An accompanying paper by Joulin and colleagues report discriminatory results in a sentiment classification task comparable to deep learning approaches while learning and predicting orders of magnitude faster[18]. *fastText* can learn text classification models on either their own embeddings or a pre-trained set (from *word2vec* for example). The *fastText* model consists of a single layer network with input of text and labels (one document may have multiple labels). The embedded words/n-grams are averaged to form the text representation later used in a linear classifier where the output is a probability distribution over the predefined classes. The representation of words/n-grams is performed with a look-up table over the vocabulary using the hashing trick[18] and is thus very fast.

  A limitation of the simplicity of both *word2vec* and *fastText* is that they require large datasets to learn generalizable embeddings. For our classification task, our labeled dataset of 542 abstracts (143599 words, 4054 unique) is too small for embeddings to be predictive – the model performs comparable to random chance (i.e. each class is predicted with ⅓ probability). Instead, a much larger corpus is necessary to learn embeddings.

*Experimental Design/ Performance Evaluation*

To assess the utility of specific and general embeddings, we assessed predictive performance in a manner identical to Surkis et al.[2]. We used 5-fold cross validation where 5 models are created, and each is trained by leaving out a different fifth of the labeled data to later validate upon. We report averaged performance across 5 folds. The cross validation folds are consistent across all models described. *fastText* is employed in its *pretrainedVectors* option (which uses a set of embeddings supplied and simply builds the classifier on top of the given embeddings), trained with labels from 4 folds and tested on the 5th fold. In all cases, we use area under the curve (AUC) as the measure of classification performance.

**Results**

*Baseline Models*

The four baseline models are trained and tested in a 5-fold cross validation experiment. The mean [range] AUC for each class T0, T1T2, and T3T4 are described in Table I.

**Table 1.** Predictive performance of the four baseline models.

| Model | AUC T0 (n = 281) | AUC T1T2 (n = 109) | AUC T3T4 (n = 152) |
|---|---|---|---|
| Naive Bayes | 95.0% [93.2, 96.8] | 81.6% [72.4, 86.6] | 90.6% [84.4, 93.9] |
| Support Vector Machines | 94.8% [93.1, 96.9] | 73.2% [67.8, 82.4] | 87.3% [78.8, 91.7] |
| Random Forest | 94.8% [92.4, 96.7] | **86.1%** [79.1, 91.0] | 88.5% [84.1, 90.9] |
| Bayesian Logistic Regression | **96.1%** [95.6, 96.9] | 85.7% [81.0, 91.0] | **92.6%** [85.5, 96.5] |

Table 1 describes respectable performance values especially considering the bag-of-words assumption made by each of these models and the relatively small sample size. Bayesian logistic regression (BLR) is the best performing baseline model with the highest performance in both classes T0 and T3T4, but marginally trails behind random forest in class T1T2. This result is consistent with prior results[23].

*General Word Embeddings*

Some descriptive information about each set of vectors and their mean [range] AUC within 5-fold cross validation are reported in Table II. Mean AUC for class T0 ranged from 94–96%, for T1T2 81–88%, and for T3T4 86–92%. In general, embeddings learnt on corpora more relevant to biomedical research performed better with the exception of the Wikipedia set which used *fastText's skip-gram model* which incorporates character-level information.

Interestingly, the *fastText* embeddings learnt on the entire English Wikipedia work very well in this task. The diverse topics covered by Wikipedia may provide a rich corpus from which to learn text semantics. In addition, Wikipedia likely contains documents related to biomedical research such that the vocabulary is not as limited compared to Freebase and GoogleNews corpora. Performance using the GoogleNews embeddings is comparable to Pubmed and Pubmed+Wiki. This result suggests that *learning embeddings in a domain-specific corpus is not a requirement for success in these tasks*.

*Specific Word Embeddings*

The mean [range] AUC results for each of the three custom-made embeddings in a 5-fold cross validation experiment are described in Table III along with the best baseline model (BLR) and off-the-shelf model (Wiki) for reference. Also included in Table III is a qualitative assessment of cost of each model and an approximate runtime. CBOW performs slightly worse than Wiki however, as previously reported for internal biomedical tasks[7,8], skip-gram outperformed CBOW (1-2%). Surprisingly, *fastText*'s skip-gram does not improve on *word2vec*'s skip-gram.

Parallel coordinate plots of ranked predicted probabilities. Each of the eight models is one vertical axis where one line is one article across the different models. The upper panel contains predictions of class T1T2 in fold 2, and the lower panel class T3T4 in fold 1.

**Table 2.** Descriptive details and predictive performance of five off-the-shelf word embeddings.

| Name | Data source(s) | Creator | Unique tokens | Model | Optimization | Dimensions | AUC T0 (n = 281) | AUC T1T2 (n = 109) | AUC T3T4 (n = 152) |
|---|---|---|---|---|---|---|---|---|---|
| Freebase | Freebase | *word2vec*[5] | 1.4 M | skip-gram | | 1000 | 94.1% [93.4, 96.0] | 81.2% [78.2, 86.3] | 86.3% [83.2, 89.7] |
| Google News | Google news | *word2vec*[6] | 3.0 M | CBOW | negative sampling | 300 | 94.7% [94.0, 96.1] | 85.9% [83.9, 87.2] | 91.3% [87.4, 95.3] |
| PubMed | PubMed+ PMC | BioNLP[5,14,15] | 4.1 M | skip-gram | hierarchical softmax | 200 | 94.6% [93.7, 96.0] | 86.0% [83.9, 87.2] | 91.1% [88.2, 95.1] |
| PubMed + Wiki | PubMed+ PMC+ Wikipedia | BioNLP[5,14,15] | 5.4 M | skip-gram | hierarchical softmax | 200 | 94.6% [93.7, 96.2] | 86.4% [83.5, 88.1] | 91.1% [88.2, 94.2] |
| Wiki | English Wikipedia | *fastText*[14] | 2.5 M | *fastText* skip-gram | negative sampling | 300 | **95.5%** [94.1, 96.4] | **88.1%** [85.3, 91.1] | **92.2%** [90.0, 94.7] |

**Table 3.** Predictive performance of three specific word embeddings models compared to the best performing benchmark and off-the-shelf models.

| | Name | Model | AUC T0 | AUC T1/T2 | AUC T3/T4 | Cost | Runtime |
|---|---|---|---|---|---|---|---|
| The best benchmark (Table 1). | BLR | Bayesian Logistic Regression | **96.1%** [95.6, 96.9] | 85.7% [81.0, 91.0] | 92.6% [85.5, 96.5] | cheap | < 5 minutes |
| The best off-the-shelf (Table 2). | Wiki | *fastText* skip-gram | 95.5% [94.1, 96.4] | 88.1% [85.3, 91.1] | 92.2% [90.0, 94.7] | cheap | < 5 minutes |
| custom | CBOW | CBOW | 94.2% [91.8, 95.9] | 87.6% [82.4, 91.9] | 90.2% [87.7, 93.7] | expensive | ~ 2 hours (28 cores) |
| | Skip | skip-gram | 95.5% [94.0, 96.3] | **88.6%** [84.7, 91.9] | **92.8%** [89.6, 95.4] | very expensive | ~ 9 hours (28 cores) |
| | fastText | skip-gram | 95.4% [93.9, 96.5] | 88.1% [84.1, 92.0] | 92.7% [89.9, 95.6] | very expensive | ~ 10 hours (28 cores) |

*Model Comparison (Parallel Coordinate Plots)*

We analyze individual documents at the micro level. We fix the classifier and use parallel coordinate plots to explore how embeddings change the ranked probabilities. Specifically, each embedding based *fastText* model is compared by ranking their predicted probabilities in a given fold (n ~ 108). Ranks are used to prevent poorly calibrated probabilities skewing the results but have their own drawbacks, namely a large difference in probability can be shrunk to a rank difference of 1 whereas a cluster of similar, but unequal, predicted probabilities will be spread over a large range of ranks. Ranks for each model are visualized in parallel coordinate plots where the ranks of each model is one vertical axis and individual articles are connected by lines. It would not be valid to consider anything other than one plot for each combination of class and fold. From the fifteen combinations of class and folds, two are presented in Figure 1. In this figure, color hue (either blue or red) is used to highlight articles that are known to be (blue) or not to be (red) in each class and color saturation depicts rank for the *word2vec* skip-gram model intended to ease the reader's task of tracking one line left to right.

*Error Analysis*

Nine articles are labeled in Figure 1 labeled into four groups, A, B, C, and D roughly describing four types of error.

A. "A" Articles
- Article A1: This article describes a meta-analysis and is thus labeled T3T4. However, the models moderately to strongly predict its class as T1T2. A human coder would observe the term 'meta-analysis' and correctly label T3T4.
- Article A2: This article describes a study in mice and is thus labeled T0. However, the models moderately to strongly predict its class as T3T4. A human coder would correctly code T0 based solely on the mention of mice.
- Analysis: Unfortunately, it appears the models do not learn this kind of rule based pattern where any mention of 'meta-analysis' or 'mice' determines an article's class. Biomedical research language can be specific and jargonized where certain words or phrases can only have one meaning. Interestingly, the specific skip-gram models perform the worst in this regard, whereas CBOW performs quite well. The large difference is likely caused by the differing model architectures. Performance may be improved in these articles if the model is exposed to more animal studies during training or if documents composed of simple sentences describing animal studies (or meta-analyses) are added to supplement the training data.

B. "B" Articles
- Article B1: This article describes the secondary use of data collected from a previous randomized controlled trial and thus labeled T3T4. However, it's abstract describes the trials and are subsequently predicted very strongly as T1T2 by every model considered.
- Article B2: This article also describes secondary use of data from two prior trials and is thus T3T4. However, the language in the abstract is indicative of a T1T2 trial and is thus very highly predicted as T1T2.
- Analysis: The word embedding based models fail in these two cases, likely due to the inability of any model tested to incorporate relative time to distinguish previous work from current contributions. Specific techniques would have to developed to differentiate these articles and a solution is not straightforward. Unfortunately, biomedical research is always written in the past tense which could complicate this task.

C. "C" Articles
- Article C1: This article describes a small randomized trial of an intervention and is thus labeled T1T2. However, the study focuses on a specific population (low-income African-American women) and is related to drug use, an issue more often addressed with T3T4 research.
- Article C2: This article describes a study investigating variation between imaging and pathology assessments and is labeled T1T2. However, the authors use data collected in a randomized controlled trial and describe the trial and its participants suggesting that it may be described as T3T4.
- Article C3: This article describes a small observational study of the safety of a rescue intervention in neonates and is labeled T1T2. However, the authors use retrospective chart review and compare outcomes, typical of T3T4 articles.
- Analysis: Although, as Figure 1 describes, the ranks are variable, each model incorrectly predicts T3T4 higher than T1T2 for each of these three articles. These three are examples of work that is atypical and could be described as spanning these two classes.

D. "D" Articles
- Article D1: This article describes a study on adoption of electronic health records with meaningful use incentives and is labeled T3T4. Every model except Freebase predicts T1T2 with very low rank. The probability of T1T2 is substantially higher (28% vs. <0.01%).
- Article D2: This article describes a study using a database of previously collected nurse surveys and is labeled T3T4. The majority of models correctly predict T3T4 with high probability except Freebase which ranks it drastically lower. The probability is lower (55% vs. 65–85%) but Freebase also predicts other articles higher than this particular one.

- Analysis: Consistent with Table II, Figure 1 describes that of the models tested, the model using the Freebase embeddings are most likely to mislabel. D1 and D2 highlight the limitations of the Freebase embeddings. D1 is a known T3T4 predicted as unlikely to be T1T2 by all models except Freebase which predicts it rank 30, 44 higher than the next model, GoogleNews. D2 is a known positive T3T4 article predicted highly likely to be T3T4 by every model except Freebase which predicts rank 53, 38 lower than the next model, PubMed. The poor performance of Freebase is likely due to the content used to learn the embeddings. Freebase was an open, collaboratively constructed database used predominantly for knowledge mining and may not represent a sufficiently diverse corpus to adequately learn semantics in biomedical language.

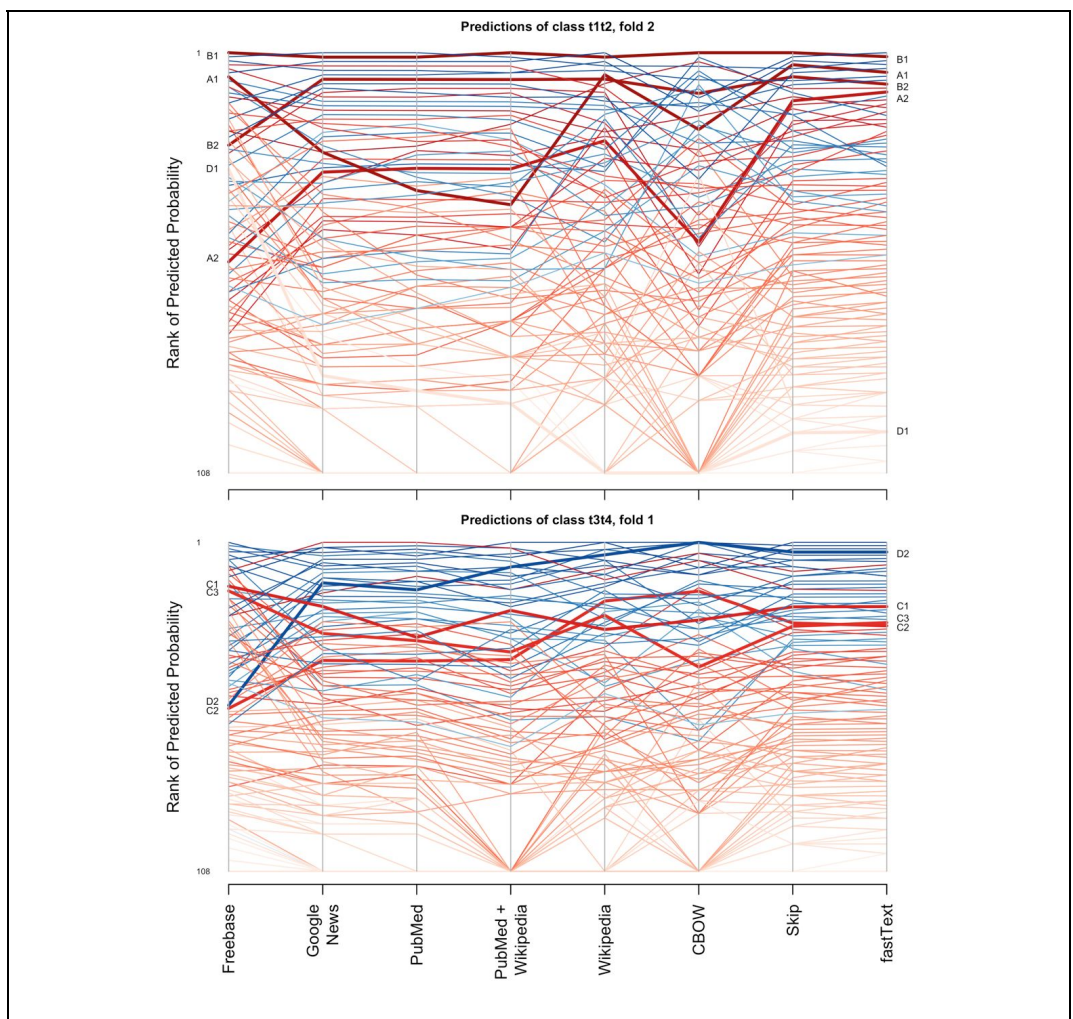

**Figure 1.** Parallel coordinate plots of ranked predicted probabilities. Each of the eight models is one vertical axis where one line is one article across the different models. The upper panel contains predictions of class T1T2 in fold 2, and the lower panel class T3T4 in fold 1.

**Discussion**

*General vs specific word embeddings*

For this task, domain-specific word embeddings using a parameter optimized *word2vec* did not drastically improve predictive performance over a set of off-the-shelf vectors. We cannot say definitely why specific word embeddings did not improve classification performance but we hypothesize several possibilities.

1) Each class is fairly homogeneous. Thus the signals are strong in each class and defined by a relatively small collection of words. These words are well represented in general word embeddings and thus specific word embeddings have limited value.

2) The specificity of language in a scientific discipline such as medicine is fairly precise and specific and consistently used in each class. This precision makes certain words unique with little overlap to other concepts. For example, a meta-analysis does not have many equivalents and thus creating a word embeddings doesn't semantically make its representation any more specific.

3) The limits of the representation of the documents and the ceiling of performance. For each class, we may have hit the ceiling of how the documents are represented and it will require more robust representations to get additional performance.

Off-the-shelf embeddings have broad utility and assurance that they are internally validated. A restricted vocabulary does not appear to be a critical issue as in the case in this work, *fastText* can impute a vector for a missing word by its character n-grams. In addition, Facebook Artificial Intelligence Research recently released an update to *fastText* (*fastText.zip*) that drastically reduces the runtime, memory usage and size of the embedding file [24] making natural language processing with embeddings even more accessible. However, for performance critical work or a very specialized domain specific embeddings may prove advantageous.

*Bayesian Logistic Regression*

Surprisingly and somewhat unexpectedly, in this task, classical models such as BLR perform the best in the T0 class. Word embeddings do not always improve classification performance. Here also we cannot say definitively why the word embeddings (general or specific) did not improve classification performance. However, we hypothesize at least two possibilities. There are some tokens and words that are highly indicative of the T0 class and additional word embeddings introduce noise and confuse the classifier. Additionally, word embeddings rely on the consistency of natural language to learn semantic meaning. Abstracts are often terse and dynamic and word embeddings algorithms may not excel in learning general patterns of language in such a restricted format.

*Computation Cost*

For these tasks, general pre-computed word embeddings are surprisingly robust. They provide near state of the art performance at very little cost. However, if the use case dictates maximizing performance and time allows, then it is more than reasonable to spend the computational time to build specific word embeddings. Outside of empirical experimentation, it is difficult to predict whether a specific embedding will help.

**Limitations**

The scope of our findings are limited to our 3 classification tasks in MEDLINE. It is possible to construct tasks and examples where specific embeddings may incur much greater performance improvements and is a point of further experimentation and exploration.

The sample itself is fairly small but also typical of tasks in the biomedical domain. Labeling is expensive due to the requisite domain knowledge needed and variable due to the complexity of the labeling protocols. Additional labeled data is always warranted and can help determine whether the model has plateaued in performance. In these cases, word embeddings can have a massive influence on learning in small sample. Additional work would explore whether performance of the system has plateaued and how little sample is needed for stable classification.

The general word embeddings may suffer from vocabulary mismatch due to varying preprocessing (for example, punctuation/accents, grammatical case, digits, stemming etc.) between the off-the-shelf embeddings and our labeled dataset. This result may worsen performance as certain tokens may not match and thus are considered out of vocabulary and have no corresponding vector. For example, contractions such as "*can't*" are split into "*can t*" by our preprocessing.

**Conclusions**

This work introduced a set of translational classification tasks. We demonstrated that specific word embeddings can further push the state of the art in maximizing discriminatory performance in classification tasks where sufficient data is available. However, in the absence of sufficient domain-specific data, conventional classifiers, those without word embeddings, perform well and classifiers using general, off-the-shelf, word embeddings also perform quite well with little additional runtime and may, in fact, prove to be more generalizable given they tend to encompass a vast corpus of text. In conclusion, experimentation across different settings and tasks is warranted to obtain maximal classification performance.

**Acknowledgement**

We acknowledge NYU Center for Health Innovation and Delivery Science, NYU Medical Center IT, and NYU High Performance Computing.